\documentclass{article}
\PassOptionsToPackage{numbers, compress}{natbib}
\usepackage[preprint]{academic_preprint}

\usepackage[T1]{fontenc}
\usepackage[utf8]{inputenc}

\usepackage{xspace}
\usepackage[dvipsnames]{xcolor} 
\usepackage{graphicx}
\usepackage{amssymb}
\usepackage{amsmath}
\usepackage{mathtools}
\usepackage{booktabs}
\usepackage{url}
\usepackage{subcaption}
\usepackage{hyperref} 
\usepackage{enumitem} 
\usepackage{multirow}
\usepackage[capitalize,nameinlink]{cleveref}
\usepackage{makecell} 
\usepackage{algorithm2e}
\crefname{algocf}{algorithm}{algorithms}
\Crefname{algocf}{Algorithm}{Algorithms}
\newcounter{supplement}
\crefname{supplement}{Supplement}{Supplements}
\Crefname{supplement}{Supplement}{Supplements}

\usepackage{ulem} 
\usepackage{wrapfig}

\AtBeginDocument{
  \providecommand\BibTeX{{\rmfamily B\kern-.05em{\scshape i}\kern-.025em b}\TeX}
}

\newcommand{\method}{{\tt YoursProtoP}\xspace} 
\newcommand{\pipnet}{PIP-Net\xspace}

\hypersetup{
    colorlinks=true,
    linkcolor=blue,
    citecolor=blue,
    urlcolor=blue
}

\title{Personalized Interpretability - Interactive Alignment of Prototypical Parts Networks}

\author{
  \begin{tabular}[t]{@{}c@{\hspace{3em}}c@{\hspace{3em}}c@{}}
    \textbf{Tomasz Michalski}$^{1,2,\dag}$ & \textbf{Adam Wróbel}$^{1,2}$ & \textbf{Andrea Bontempelli}$^{3}$ \\
    \rm Jagiellonian University & \rm Jagiellonian University & \rm University of Trento \\[3ex]
    \textbf{Jakub Luśtyk}$^{4}$ & \textbf{Mikolaj Kniejski}$^{5}$ & \textbf{Stefano Teso}$^{3,6}$ \\
    \rm Transmission Dynamics & \rm University of Warsaw & \rm University of Trento \\ [3ex]
    \textbf{Andrea Passerini}$^{3}$ & \textbf{Bartosz Zieliński}$^{1}$ & \textbf{Dawid Rymarczyk}$^{1,7}$ \\
    \rm University of Trento & \rm Jagiellonian University & \rm Jagiellonian University \\
    & & \rm Ardigen SA
  \end{tabular}
}
\begin{document}
\maketitle
\renewcommand{\thefootnote}{\fnsymbol{footnote}}
\footnotetext{$^{1}$Faculty of Mathematics and Computer Science, Jagiellonian University, Kraków, Poland.}
\footnotetext{$^{2}$Doctoral School of Exact and Natural Sciences, Jagiellonian University, Kraków, Poland.}
\footnotetext{$^{3}$Department of Information Engineering and Computer Science, University of Trento, Italy.}
\footnotetext{$^{4}$Transmission Dynamics Poland sp.\,z\,o.\,o., Henryka Pachońskiego 9/K-22, 31-223 Kraków, Poland}
\footnotetext{$^{5}$Faculty of Psychology, University of Warsaw, Poland.}
\footnotetext{$^{6}$Center for Mind/Brain Sciences, University of Trento, Italy.}
\footnotetext{$^{7}$Ardigen SA, Leona Henryka Sternbacha 1, 30-394 Kraków, Poland.}
\footnotetext{$^{\dag}$Corresponding author: tomasz.michalski@doctoral.uj.edu.pl}

\begin{abstract}
Concept-based interpretable neural networks have gained significant attention due to their intuitive and easy-to-understand explanations based on case-based reasoning, such as ``this bird looks like those sparrows".
However, a major limitation is that these explanations may not always be comprehensible to users due to concept inconsistency, where multiple visual features are inappropriately mixed (e.g., a bird's head and wings treated as a single concept). This inconsistency breaks the alignment between model reasoning and human understanding.
Furthermore, users have specific preferences for how concepts should look, yet current approaches provide no mechanism for incorporating their feedback.
To address these issues, we introduce \method, a novel interactive strategy that enables the personalization of prototypical parts---the visual concepts used by the model---according to user needs. By incorporating user supervision, \method adapts and splits concepts used for both prediction and explanation to better match the user's preferences and understanding.
Through experiments on both the synthetic FunnyBirds dataset and a real-world scenario using the CUB, CARS, and PETS datasets in a comprehensive user study, we demonstrate the effectiveness of \method in achieving concept consistency without compromising the accuracy of the model.
\end{abstract}

\section{Introduction}

Despite the remarkable success of deep learning methods across various domains, a significant challenge remains in making these powerful but opaque models interpretable to humans~\citep{rudin2019stop}. To tackle this, the field of Explainable Artificial Intelligence (XAI) has emerged~\citep{ribeiro2016should}. Early efforts in XAI for image recognition relied on heatmap-based methods, such as saliency maps~\citep{bach2015pixel,selvaraju2017grad}. These methods aim to highlight important pixels from the model's perspective, but their reliability has been questioned~\citep{adebayo2018sanity}.

To overcome these limitations, explainable-by-design neural architectures have been developed, including methods such as Concept Bottleneck Models (CBMs)~\citep{koh2020concept} and Prototypical-Parts Networks (e.g., PIP-Net)~\citep{nauta2023pip}. These methods provide explanations in the form of high-level concepts, following the principle ``this looks like that''~\citep{chen2019looks}. However, users often struggle to identify their meaning because these concepts can be vague~\citep{kim2022hive}. There were attempts to address this issue through improved visualizations~\citep{ma2023looks} or concept decomposition into low-level vision features~\citep{pach2024lucidppn}. However, the mixing of multiple features can still occur, which is a common issue in deep neural networks~\citep{stammer2022interactive}.

\begin{figure}
    \centering
    \includegraphics[width=0.8\textwidth]{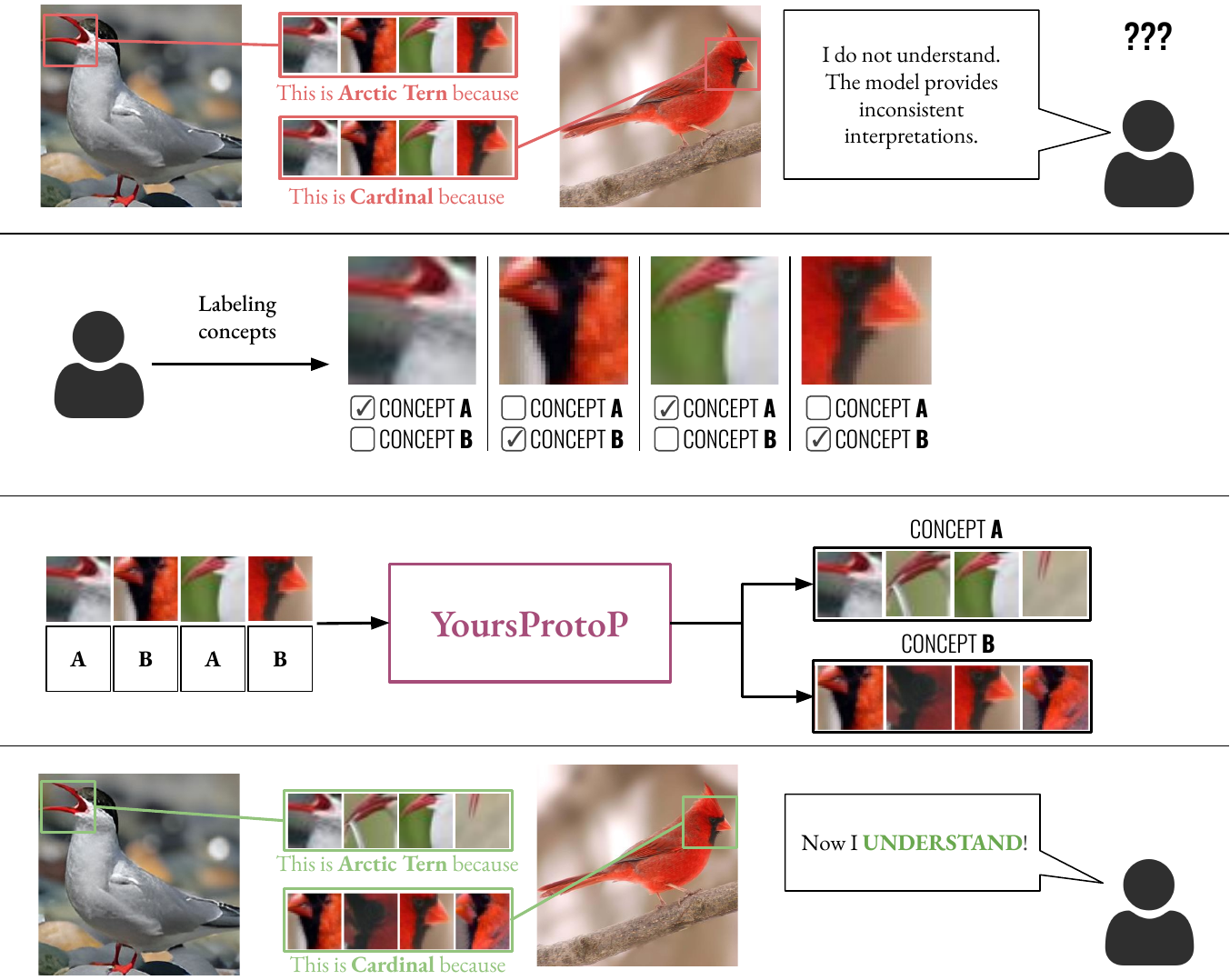}
    \caption{The core idea behind \method is to enable user-guided splitting of \textit{inconsistent} prototypes. The user begins by selecting a prototypical part they perceive as inconsistent. Then they annotate several patches within this prototype, assigning them two distinct concepts (A or B). These labeled patches are then incorporated into the training pipeline, where \method dynamically adapts the architecture to refine the prototype representations according to the user’s intent. }
    \label{fig:teaser}
    \vspace{-1.5em}
\end{figure}

To address this challenge, we propose a novel approach that integrates user feedback to improve the quality of interpretations of concept-based models. While previous works focused on removing confounders and unwanted concepts~\citep{bontempelli2022concept,kim2024constructing,xuan2024slim}, we propose \method (see~\Cref{fig:teaser}) that goes a step further. It leverages users' feedback supported with an automated prototype\footnote{Prototype and prototypical part are used interchangeably.} selection strategy to identify inconsistent concepts and split them through a simple yet effective fine-tuning procedure, without compromising the model's accuracy. Our automatic selection methodology identifies the most inconsistent prototypes by analyzing feature similarity patterns within the model's representation space, significantly reducing the user's corrective supervision while delivering explanations tailored to user needs. We thoroughly test the splitting process with the FunnyBirds~\citep{hesse2023funnybirds} dataset, which enabled the development and evaluation of concept-based models without requiring extensive user studies during the design phase. Furthermore, we validate \method's effectiveness in real-world scenarios using the CUB~\citep{welinder2010caltech}, CARS~\citep{krause20133d}, and PETS~\citep{parkhi2012cats} datasets, demonstrating that our automated selection strategy identifies and splits inconsistent prototypes while maintaining classification performance.

Our contributions can be summarized as follows:
\begin{itemize}[leftmargin=1.25em]
\item We propose \method, a concept-based model for personalized interpretability to achieve consistent concepts without compromising model accuracy.
\item We develop an automated prototype selection strategy that identifies inconsistent concepts by analyzing feature similarity patterns within the model's representation space, significantly reducing the burden of manual identification.
\item We extensively evaluate \method on synthetic and real world datasets.
\end{itemize}

\section{Related Works}
\label{sec:related-work}

\paragraph{Concept-based models.} Several concept-based models have been proposed including concept bottleneck models~\citep{koh2020concept} and Prototypical Parts Network (ProtoPNet)~\citep{chen2019looks}. The ProtoPNet classifies images by comparing them with a fixed number of prototypical parts for each class. ProtoPShare builds upon ProtoPNet~\citep{rymarczyk2020protopshare} by minimizing the explanation size through pruning of the prototypes based on their semantic similarity. To avoid the additional step of pruning, ProtoPool~\citep{rymarczyk2022interpretable} introduces a soft assignment of the prototypes to the classes. This significantly improves interpretability by reducing the number of prototypes. 
However, the prototypes learned by these models may not correlate with the user's concepts because a single prototype may represent multiple concepts. It stems from the assumption that the images of the same classes have assigned the same prototypes to them. To address this limitation, \pipnet~\citep{nauta2023pip} allows prototypes to be shared across classes, and each prototype activation adds or abstains from classification. This further reduces the number of prototypes. Moreover, \pipnet architecture, through alignment loss, makes interpretations semantically similar, which further improves interpretability. Furthermore, the visualization of the prototype on a single image, like in the mentioned methods, makes it difficult to understand the underlying concept. This limitation has been addressed by ProtoConcepts, which introduces the visualization on multiple image patches~\citep{ma2023looks}. On the other hand, LucidPPN~\citep{pach2024lucidppn} aims to decompose the prototypical part into low-level features to better understand the impact of color on the prototype. Despite these advancements, a critical limitation persists across all these approaches: they lack mechanisms for incorporating user feedback to correct concept inconsistency when a prototype erroneously combines multiple distinct visual features. Our proposed \method directly addresses this gap by enabling interactive refinement of prototypical parts through a user-guided splitting process, resulting in explanations that better align with human conceptual understanding while maintaining model accuracy.

\paragraph{Explanatory debugging.}
\method is inspired by work in explanatory debugging \citep{kulesza2015principles} and explanatory interactive learning \citep{teso2019explanatory,schramowski2020making}. The human user is involved in the training loop and revises the model by interacting via the model's explanations. Therefore, the feedback improves the model's reasoning and fixes bugs such as removing confounders and unwanted concepts~\citep{bontempelli2022concept,kim2024constructing,xuan2024slim}. Recent works involved pixel-level supervision to improve prototypes by penalizing the activation on irrelevant areas of the input~\citep{barnett2021case}. However, due to the high labour requirements of this approach, a concept-level debuggers has been developed where users are presented with activations of the prototypes on a limited number of images~\citep{bontempelli2022concept}. Here, the annotations require less effort and can be generalized across instances. Additionally, users prefer interaction through prototypes because these explanations are visual and informative~\citep{kimHelpMeHelp2023}. The \method collects concept-level supervision to align the prototypical part to the user interpretation, but does not provide classification disentanglement as in~\citep{stammer2022interactive}. Furthermore, it is an answer to concerns of wrong interpretations raised in~\citep{kim2022hive} and the demand by users to improve the interpretations of deep learning models posed in~\citep{kimHelpMeHelp2023}.

\section{Methods}
\label{sec:methods}

To make this work self-contained, we first introduce PIP-Net, an architecture on which we base our \method. Then, we describe the \method itself.

\subsection{PIP-Net}
\label{sec:pip-net}

\begin{figure}
    \centering
    \includegraphics[width=0.8\textwidth]{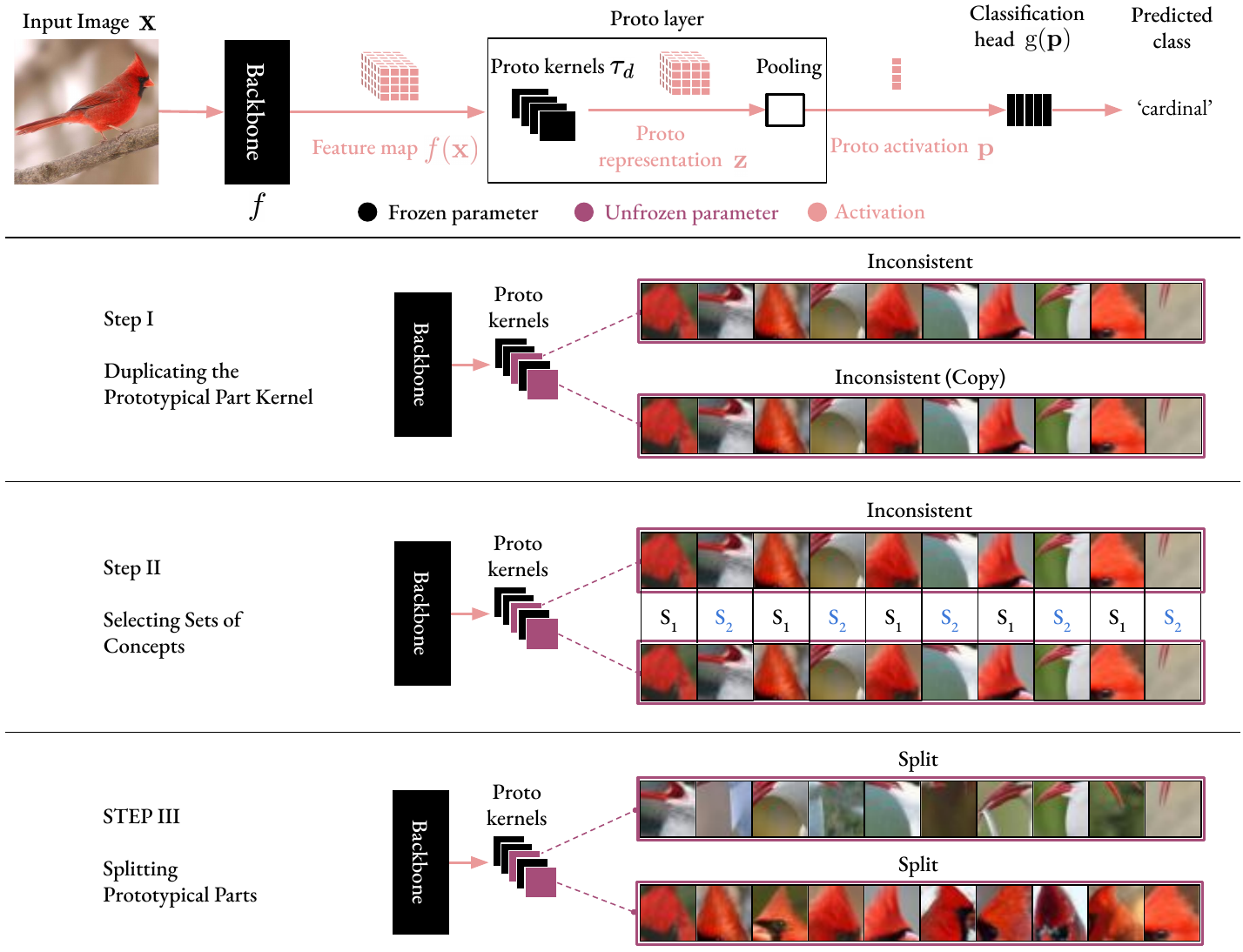}
    \caption{\method{} builds on the PIP-Net architecture presented in the top row. Following the selection of an inconsistent prototype, the entire model architecture is frozen, except for the kernel corresponding to the selected prototype. This kernel is duplicated into an additional prototype slot (Step I). Then, the method partitions inconsistent prototype patches into two distinct concept sets, $S_1$ and $S_2$ (Step II). Finally, the original and duplicated kernels are fine-tuned independently, each to a separate concept (Step III). }
    \label{fig:training}
    \vspace{-1.5em}
\end{figure}

As depicted in the top row of~\Cref{fig:training}, the PIP-Net architecture consists of a convolutional backbone $f$, prototype kernels $\tau_{d}$, pooling, and a classification head $g$.
Let $\textbf{x} \in \mathbb{R}^{H_{in} \times W_{in} \times 3}$ be an input image, and let $f: \mathbb{R}^{H_{in} \times W_{in} \times 3} \rightarrow \mathbb{R}^{H \times W \times D}$ be a backbone network for feature extraction, where $H, W$ are the spatial dimensions of the feature map and $D$ is the number of feature channels.
After obtaining the representation $\textbf{z} = f(x) \in \mathbb{R}^{H \times W \times D}$, PIP-Net applies a softmax to the third dimension $D$. Here, each spatial location $(h,w)$ corresponds to a small region in the original image, which we refer to as a \textit{patch}. The softmax operation forces each patch to belong to exactly one prototype. Each channel represents a prototypical part, and we will use those terms interchangeably. Moreover, we will use the term \textit{prototype kernel} $\tau_e$ to describe the kernel that generates the channel. Finally, the max-pooling operation over the spatial dimensions $H×W$ is used to obtain a vector of prototype activations $\textbf{p}$ defined as:
$$\textbf{p} = \left[\max_{(h,w) \in H \times W} \textbf{z}_{h,w,d}\right]_{d=1,\dots,D} \in [0,1]^D.$$
Finally, $\textbf{p}$ is passed to a sparse classification layer with non-negative weights $\boldsymbol{\Omega}\in \mathbb{R}_{\geq 0}^{D \times K}$ to obtain prediction scores $\textbf{o} = g(\textbf{p}) = \textbf{p}\boldsymbol{\Omega}$ for $K$ classes, where each element $o_k$ represents the score for class $k$.

\begin{wrapfigure}[22]{r}{0.6\textwidth}
\vspace{-13pt} 
    \centering
    \begin{subfigure}[b]{0.95\linewidth}
        \includegraphics[width=\linewidth]{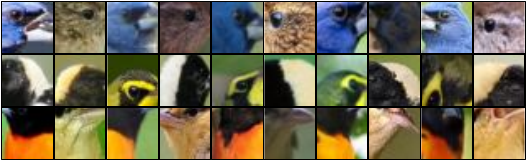}
        \caption{Examples of inconsistent prototypical parts.}
        \label{fig:entangled_prototype}
    \end{subfigure}
    
    \begin{subfigure}[b]{0.95\linewidth}
        \includegraphics[width=\linewidth]{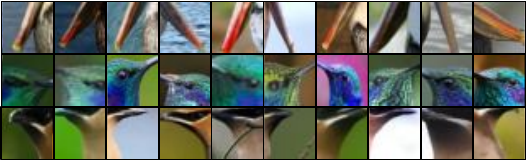}
        \caption{Examples of consistent prototypical parts.}
        \label{fig:not_entangled_prototype}
    \end{subfigure}
    
    \caption{Some of the prototypical parts can be inconsistent in terms of presented semantic concept as shown by the (a) part of this Figure. Note that each row is a separate prototypical part, and (b) part presents semantically consistent prototypical parts.}
    \label{fig:finding-entangled-prototypes}
\end{wrapfigure}

As an explanation for predicting class $k$, PIP-Net identifies the regions of the input that are responsible for the maximal activations for prototypes $d=1,\dots,D$ with weight $\boldsymbol{\Omega}_{d,k}$ greater than a given threshold. These maximal activation regions are then cut out of training images to create explanations in the form of the 10 most activated patches, such as in \Cref{fig:not_entangled_prototype}.\\
Users frequently encounter prototypical parts in PIP-Net explanations that lack conceptual clarity. A significant issue arises when a prototype exhibits \textit{inconsistency} by simultaneously representing multiple distinct visual concepts, for example, combining throat and head regions of different colors into a single representation, as illustrated in the bottom prototype of~\Cref{fig:entangled_prototype}. 
This conceptual ambiguity diminishes the interpretability of model explanations. To address this limitation, we propose \method, a novel approach that enables the separation of such inconsistent prototypical parts into their constituent concepts. The methodology for this splitting process is detailed in the following section.

\subsection{\method}\label{sec:method}
We introduce a novel method to make PIP-Net's explanations aligned with users' understanding through splitting inconsistent prototypes. We define an inconsistent prototype as $e \in \{1, \dots, D\}$ for which there exist at least two disjoint sets of patches $S_1, S_2 \subset P_e$ where $P_e$ represents all patches with highest activations for prototype $e$. In other words, we seek prototypes for which we could identify two concepts such that $|S_1| \geq Q$, $|S_2| \geq Q$, where $Q$ represents the minimal number of patches in each concept, and each set $S_i$ corresponds to a distinct visual feature. ~\Cref{fig:entangled_prototype} presents examples of such inconsistent prototypes that consist of two concepts.
\paragraph{Step 1: Duplicating the Prototype Kernel and the Corresponding Weights of the Classification Layer.}
To split a prototypical part, we need to initialize a new channel. We do that by adding a new prototype kernel $\tau_{D+1}$ that is a copy of the existing kernel $\tau_e$. As a result, the representation $\textbf{z}$ is extended to $\textbf{z}' \in \mathbb{R}^{H \times W \times (D+1)}$ by an additional channel so that for each spatial point $(h,w)$:
\begin{equation}
{z'}_{h,w,d} =
\begin{cases}
z_{h,w,d} & \text{for } d \leq D \\
z_{h,w,e} & \text{for } d = D + 1.
\end{cases}
\end{equation}
After adding a new prototype kernel, to maintain the classification performance, we also extend the weight matrix $\boldsymbol{\Omega}$ to $\boldsymbol{\Omega}' \in \mathbb{R}^{(D+1) \times K}$ so that the new prototype initially has the same weights as the original prototype $e$:
\begin{equation}\label{eq:losses}
{\boldsymbol{\Omega}'}_{d, :} =
\begin{cases}
\boldsymbol{\Omega}_{d, :} & \text{for } d \leq D \\
\boldsymbol{\Omega}_{e, :} & \text{for } d = D+1.
\end{cases}
\end{equation}

\paragraph{Step 2: Selecting Sets of Concepts.}
To split the prototype $p$ into two prototypes corresponding to two visual concepts, we need to generate three sets containing: patches exclusively of the first concept $S_1$, the second concept $S_2$, and everything except those two concepts $S_r$ (reference set). Generally, in real-world scenarios, we assume that the first or the second sets are populated either automatically or by the user, while the patches for the reference set are always obtained automatically, choosing regions with high activations of other prototypes.

\paragraph{Step 3: Splitting Prototypes.}
During training, we optimize only prototype kernels $\tau_e$ and $\tau_{D+1}$ (the duplicate of $\tau_e$) while all other network parameters remain frozen. For each element $\bold{x} \in S_1 \cup S_2 \cup S_r$ and its prototypes' activation $\bold{p}$, the splitting uses a specialized loss function:

\begin{equation}
l(\bold{x}, \bold{p}, \kappa, \alpha) =
\begin{cases}
l_{act}(p_e) & \text{if } \bold{x} \in S_1 \\
l_{act}(p_{D+1}) & \text{if } \bold{x} \in S_2 \\ \alpha [l_{deact}(p_e, \kappa) + l_{deact}(p_{D+1}, \kappa)] & \text{if } \bold{x} \in S_r
\end{cases}
\end{equation}

\noindent where:

\begin{equation}
\begin{split}
    l_{act}(x) &= -\log(x) \\
    l_{deact}(x, \kappa) &= \max\{0, -\log(1-x) - \kappa\}
\end{split}
\end{equation}

\noindent Intuitively, $l_{act}$ is designed to increase the activation of the given softmax channel and $l_{deact}$ to decrease the activation to a specified threshold $\kappa$. Since both $p_e$ and $p_{D+1}$ are components of the same softmax output vector $\bold{p}$ , increasing the activation of one reduces the activation of other, and vice versa.
This allows us to use only $l_{act}$ when the input patch $\bold{x} \in S_1$ or $\bold{x} \in S_2$. In contrast, for inputs $\bold{x} \in S_r$, we aim to jointly minimize the activations of both channels using $l_{deact}$, pushing them below the threshold $\kappa$. The parameter $\alpha$ controls the relative contribution of the deactivation loss $l_{deact}$ to the overall objective $l$. Finally, the total loss is computed by averaging the individual losses $l(\bold{x}, \bold{p}, \kappa, \alpha)$ over all inputs $\bold{x}$.

\paragraph{Finding inconsistent prototypical parts.}\label{sec:finding_entangled}
To identify inconsistent prototypical parts without manual inspection of hundreds of channels, we developed a heuristic approach that analyzes feature space clustering patterns. We construct a similarity graph where each node represents a feature vector from prototypes' patches. We then identify maximal cliques (groups of mutually similar features) using the Bron-Kerbosch algorithm. 

Prototypes containing multiple distinct cliques of sufficient size are flagged as inconsistent. We prioritize splitting prototypes that show the significant separation between concepts (highest inter-clique dissimilarity). Moreover, the cliques are used to select concepts for each prototype automatically. The detailed algorithm implementation with pseudocode is provided in the supplementary materials.

\section{Experimental Setup}\label{sec:experimental_setup}

In this section, we describe the common experimental parameters across all settings, followed by the methodology used for the synthetic dataset (FunnyBirds) and natural datasets (CUB, CARS, PETS), including a description of the user study.
We chose the ConvNext-Tiny architecture in our study because it reaches significantly higher purity rates (0.92), compared to, e.g., ResNet-50 (0.63)~\citep{nauta2023pip}. This ConvNext-Tiny backbone is trained according to parameters used in~\citep{nauta2023pip} for all the natural datasets, while we apply CUB's setup for training on the FunnyBirds dataset. Furthermore, in all experiments, for the splitting of prototypical parts, we utilize Adam optimizer with a learning rate of $1 \times 10^{-4}$ and weight decay of $10^{-4}$ with batch size 10 during splitting. To enhance generalization, we apply Gaussian noise regularization ($\sigma = 0.05$) to input features and add a small constant ($\epsilon = 10^{-8}$) for numerical stability. For our splitting loss function, we set $\alpha=2$ to prioritize the deactivation loss in~\Cref{eq:losses}, which encourages stronger separation between reference and selected concepts, while we assign $\kappa=0.1$. Based on experimental results of split prototypical parts, we set up the convergence criteria during training as either achieving $99.9\%$ per-concept accuracy or reducing the loss below $0.02$ for a consistent, patientience period. Finally, we reinitialize and finetune weights corresponding to split channels. The weights are initialized using a normal distribution with a mean and standard deviation of the other nonnegative weights in the last layer. Then, only the weights of the finetuned channels undergo a fine-tuning process for a single epoch.\\
\begin{wrapfigure}[18]{r}{0.5\textwidth}
    \vspace{-13pt}
    \centering
    \includegraphics[width=0.48\textwidth]{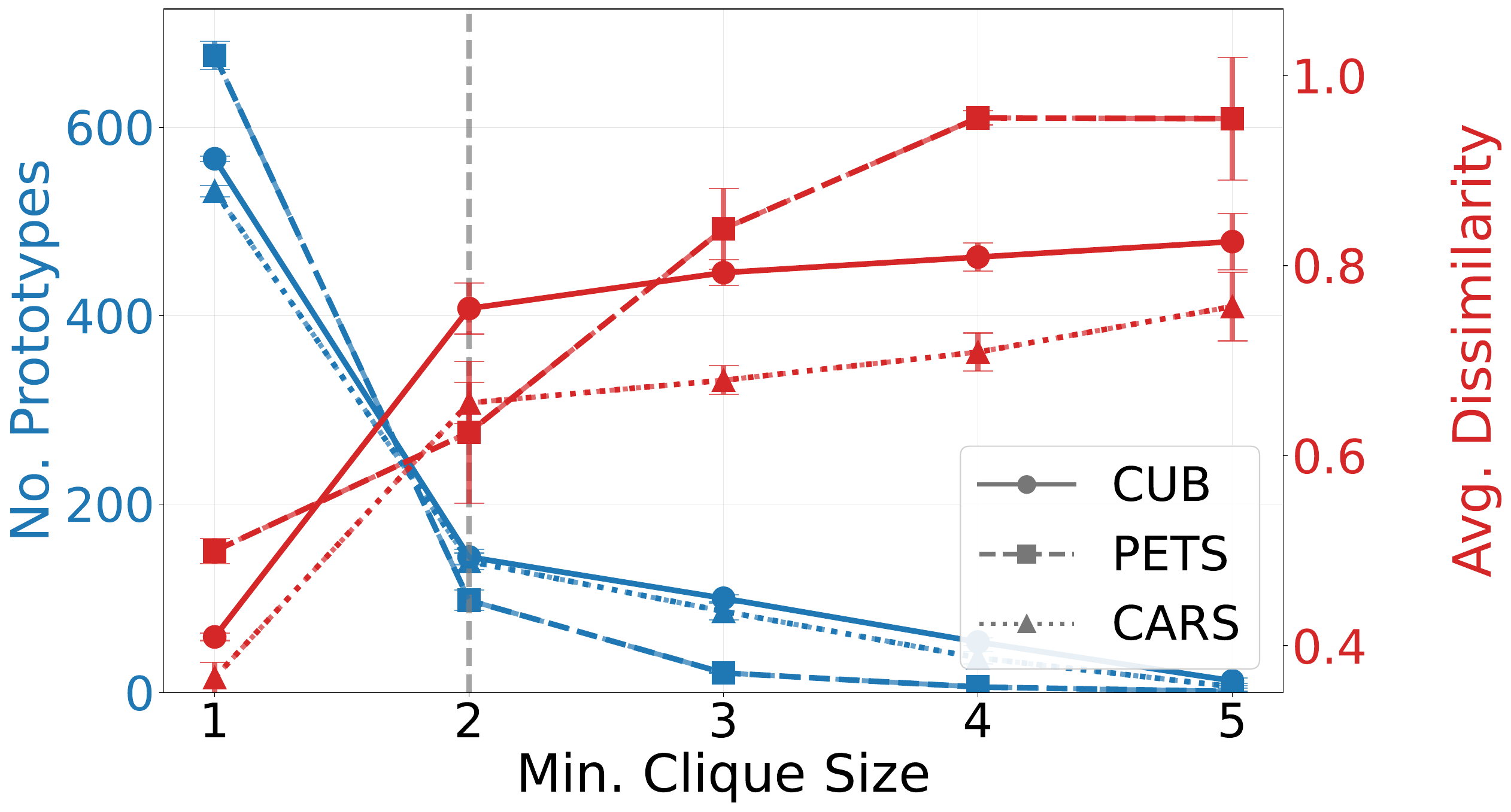}
    \caption{There is a visible breaking point at a minimal size of clique $Q=2$. As the minimal requirement for clique size increases, the number of potentially inconsistent prototypes selected by heuristics decreases, and the dissimilarity between cliques increases. The error bars indicate standard deviation (STD) for the mean obtained from 5 models for each dataset.}
    \label{fig:finding_clique}
    \vspace{-10pt}
\end{wrapfigure}
In section~\Cref{sec:method} we define an inconsistent prototypical part as containing two concepts with at least $Q$ patches each. In case of our proposed heuristics, this $Q$ value reflects the minimal clique size. To determine the optimal value for $Q$, we have performed an analysis on all of the natural datasets. We define dissimilarity between cliques as 1 minus cosine similarity between the most similar elements of the cliques. As can be seen on~\Cref{fig:finding_clique}, the dissimilarity between cliques increases dramatically as it reaches a minimum size of $Q=2$. At that point, we see a significant change in dissimilarity between the cliques. Note that dissimilarity is crucial for selecting the most inconsistent prototypical parts. Also, keeping the $Q$ low increases the chance of selecting the highest number of potentially inconsistent prototypes.

\subsection{FunnyBirds}
The FunnyBirds is a synthetic dataset consisting of 50 classes built out of 26 different bird parts. It offers several key advantages for validating our approach. First, it allows both automation of the splitting process and objective verification through the dataset's segmentation maps. Second, the concepts in this dataset have clear, exact definitions that are not open to subjective interpretations. Third, the distinct separation of concepts in FunnyBirds (with no overlapping features) simplifies prototype examination. To perform the analysis, we propose the adaptation of the \textit{purity metric} from~\citep{nauta2023pip} and we define it as \textit{pattern purity} $PP=1/k$ where $k$ is the number of distinct bird parts' combinations (\textit{patterns}) observed across a prototype's patches. In other words, we want to measure the purity of patterns observed within the 10 most activated patches. The smaller the number of patterns the higher the metric. Furthermore, based on the selected value of a minimal number of concept patches $Q$, we defined an inconsistent prototypical part for the FunnyBird's dataset as containing at least two different patterns with at least $Q$ patches each. In case of the synthetic dataset, the selection of inconsistent prototypes is performed using segmentation masks. Also, the prototypes are split starting with the most inconsistent ones, i.e, having the highest $PP$.

\subsection{Natural Datasets}
Due to the practical constraints on user engagement time, we strategically focus our efforts on the most inconsistent prototypical parts and select them with the proposed heuristics in~\Cref{sec:method}. Specifically, we choose those prototypical parts that contain groups of patches (cliques) with the highest dissimilarity between them. Then we perform automatic splitting of the channels according to the concepts assigned by the heuristics. Furthermore, since the CUB dataset contains point annotations, it allows for computation of the purity metric~\citep{nauta2023pip}. This metric determines the percentage of appearance of the most repeated concept only. Higher purity values indicate that the dominant part appears more often in the patches of a given prototype. 

\paragraph{User study.}
Since inconsistency of prototypes in natural datasets is subjective, we performed a user study to show the effectiveness of \method. For this reason, we have created a website that leads users through (1) the verification of whether a prototypical part is inconsistent, (2) the selection of concepts for splitting, and (3) the assessment of the consistency of the newly created prototypical parts. The user study is performed for each dataset (CUB, CARS, and PETS) separately. In each case, we select the 10 prototypes with the highest dissimilarity between the most similar elements within their cliques. At first, users have to decide if a presented prototype is inconsistent, and if they decide it is, they are asked to label its patches as either \textit{Concept A}, \textit{Concept B}, or \textit{Something Else}. Then, the \method splits the prototypical parts online according to users' feedback. Finally, users are asked if the newly created prototypical parts are more consistent than before the split. Note that at this stage, users have to evaluate two sets of patches for each split prototype. Images presenting the user study are shown in the supplementary materials.
All studies were conducted using the Prolific platform. To ensure statistically significant results, we recruited 18 participants for each study, for a total of 54 participants across all three datasets. The median age was 34.5 with an interquartile range of 14.75 years, while the gender ratio was 53.7\% male to 46.3\% female (29 males, 25 females).

\section{Results \& Analysis}
Our experimental evaluation follows a two-phase approach. In the first phase, we validate our prototype splitting methodology in the controlled environment of the FunnyBirds synthetic dataset. This controlled setting allows us to establish the fundamental effectiveness of our approach. In the second phase, we extend our methodology to natural images to demonstrate the real-world application of \method. 
Our evaluation examines three key research questions: (1) Is it possible to split prototypical parts into distinct conceptual representations? (2) What is the alignment between users and automatic identification of inconsistent prototypical parts (3) Does the quality of split prototypical parts increase with \method? Furthermore, we provide statistics on the user's effort required in the process of concept alignment with \method and their agreement with the concepts selected by the heruistics.

\subsection{Is it possible to split prototypical parts in the synthetic dataset?}

\begin{wrapfigure}[22]{r}{0.5\textwidth}
    \vspace{-15pt}
    \centering
    \begin{minipage}[t]{0.48\textwidth}
        \centering
        \includegraphics[width=\linewidth]{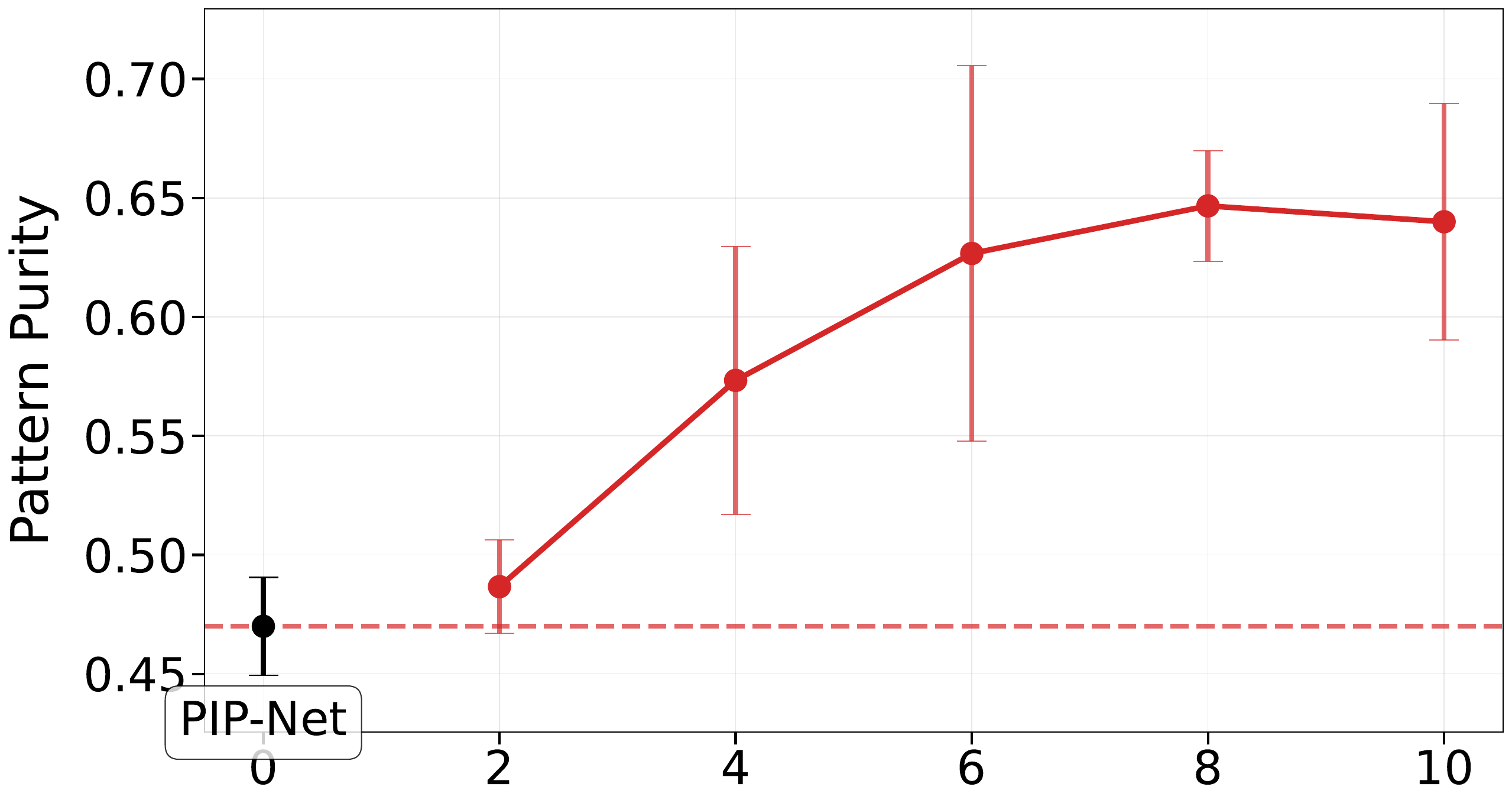}
    \end{minipage}
    \hfill
    \begin{minipage}[t]{0.48\textwidth}
        \centering
        \includegraphics[width=\linewidth]{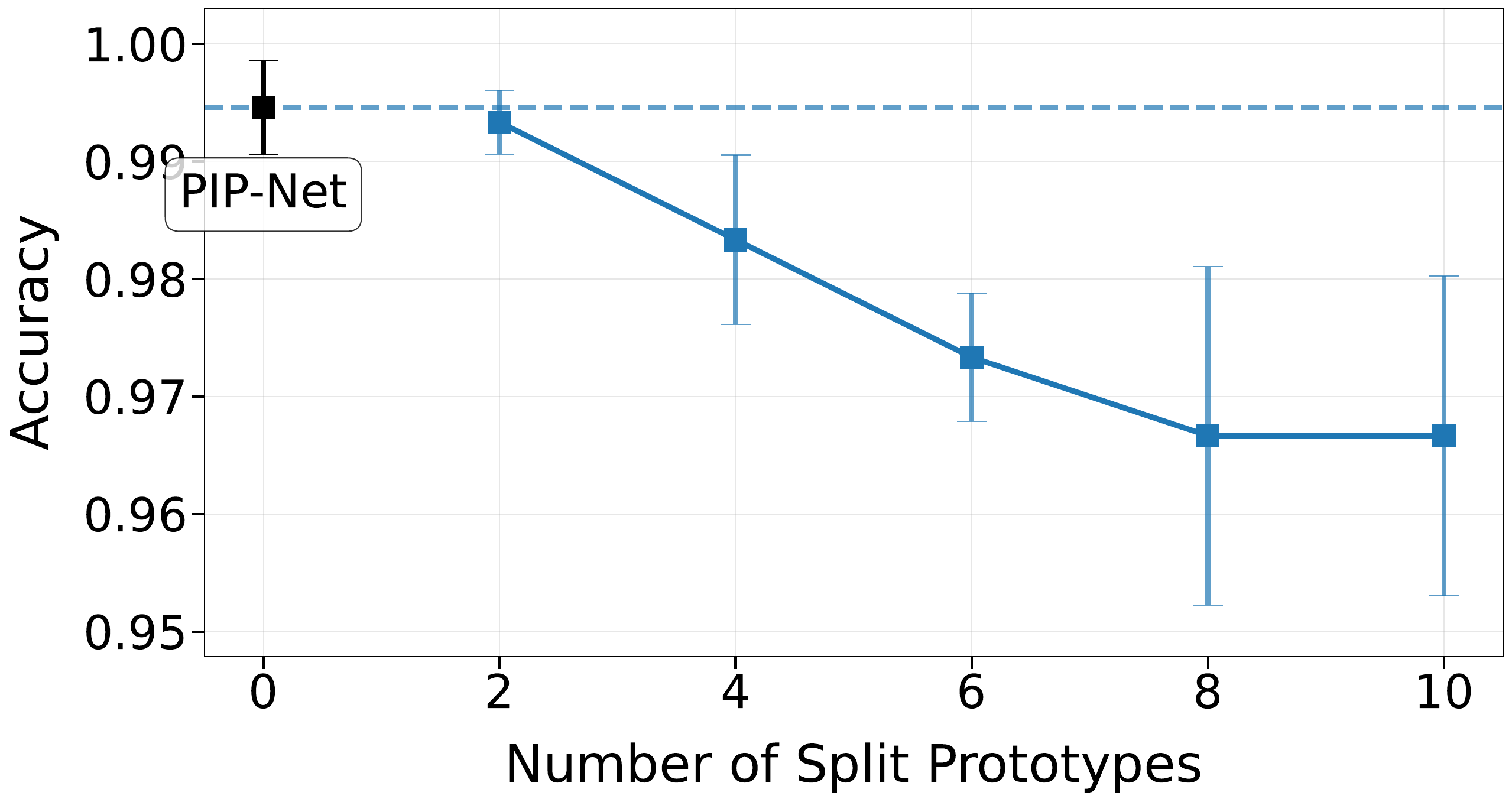}
    \end{minipage}
    \caption{Increase in $PP$ (top) and decrease of model's accuracy (bottom) for FunnyBirds. The error bars indicate the STD for the mean over 5 different models.}
    \label{fig:app_and_acc}
    \vspace{+30pt}
\end{wrapfigure}

We begin our analysis by demonstrating the effectiveness of \method using a synthetic dataset.
The top section of the~\Cref{fig:fb_cub_splitting} presents a successful split of an inconsistent prototype into two consistent ones: a red wing with a leg and a leg only. In the case of FunnyBirds, this channel is selected using segmentation masks. Then, through our \method, it is split into two separate, conceptually consistent concepts: the co-occurrence of a red wing and a leg, and a leg only.
Following the visual presentation of effectiveness on \method, now we would like to examine the influence of multiple splits with YoursProtoP on the model's performance. As shown in the top plot of~\Cref{fig:app_and_acc}, the $PP$ increase is more pronounced at the beginning of the consecutive splits. This value is computed only for the selected and split prototypical parts. Furthermore, the lower plot~\Cref{fig:app_and_acc} shows that when the number of split prototypes increases, the accuracy of the models decreases.

\subsection{Is it possible to split prototypical parts in the natural dataset?}\label{sec:results_natural}

\begin{wraptable}[16]{r}{0.5\textwidth}
\vspace{-13pt}
\caption{\method achieves better Purity than PIP-Net on CUB after performing splitting of inconsistent prototypes. Comparison of Test Accuracy and Average Purity with other architectures. (R - ResNet50, C - ConvNext-Tiny).}
\centering
\begin{tabular}{lcc}
\toprule
\textbf{Method} & \textbf{Test Acc.} $\uparrow$ & \textbf{Purity} $\uparrow$ \\
\midrule
ProtoPNet R& 79.2 & 0.44 $\pm$ 0.21 \\
ProtoTree R  & 82.2 $\pm$ 0.7 & 0.13 $\pm$ 0.14 \\
ProtoPShare R  & 74.7 & 0.43 $\pm$ 0.21 \\
ProtoPool R  & \textbf{85.5 $\pm$ 0.1} & 0.35 $\pm$ 0.20 \\
PIP-Net R  & 82.0 $\pm$ 0.3 & 0.63 $\pm$ 0.25 \\
\midrule
PIP-Net C & 84.3 $\pm$ 1.0 & 0.84 $\pm$ 0.11 \\
\method C & 84.3 $\pm$ 1.0 & \textbf{0.90 $\pm$ 0.10} \\
\bottomrule
\end{tabular}

\label{tab:accuracy_purity}
\end{wraptable}

\begin{figure}[h]
    \centering 
    \includegraphics[width=0.8\linewidth]{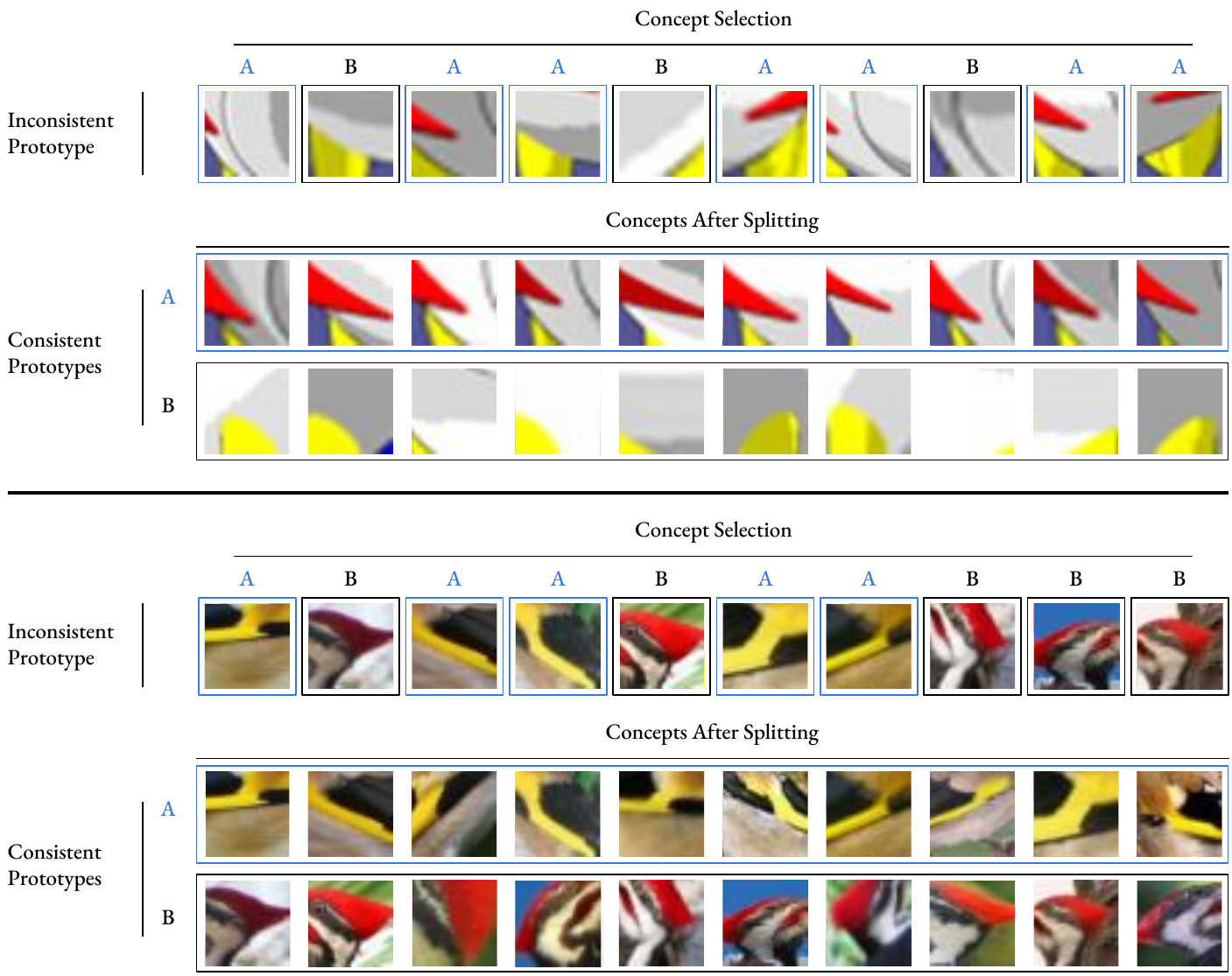}
    \caption{Examples of splitting of inconsistent prototypes in FunnyBirds (top) and CUB (bottom).}
    \label{fig:fb_cub_splitting}
    \vspace{-1.5em}
\end{figure}

After validating our method on the synthetic dataset, we extend our approach to natural images. In the bottom part of~\Cref{fig:fb_cub_splitting} we present an example of the splitting of the inconsistent prototypical part. The prototypical part confuses the head with the wing of the bird. However, it is correctly split into these two separate concepts. More visual examples are shown in the supplementary materials.
The splitting of the selected 10 prototypical parts by the heuristics results in a significant increase in purity of these channels (0.84 to .90) while the accuracy is maintained, as shown in~\cref{tab:accuracy_purity}. The \method stands out compared to other prototypical parts networks in terms of accuracy and purity metrics. 

\subsection{How does the automatic splitting align with the user's judgment?}
\Cref{tab:user_study} presents results from user studies conducted across the three natural datasets. The data demonstrates that participants identified a significant majority of the prototypical parts as inconsistent: 91.7\% for CUB, 86.1\% for CARS, and 70.6\% for PETS. Following the application of our splitting method, users confirmed substantial improvements in prototype consistency, with 88.3\%, 84.2\%, and 80.4\% of the split prototypes rated as more consistent for the respective datasets. The data also reveals important patterns in concept alignment. For the CUB and CARS datasets, we observed high agreement in concept selection both among users (89\% and 87\% respectively) and between users and our proposed heuristics (92\% and 78\%). However, the PETS dataset demonstrated notably lower agreement rates (50-54\%). This discrepancy likely stems from the potential presence of more than two major concepts within the selected PETS prototypes, making binary concept assignment more challenging and introducing greater subjectivity in the classification process. Regarding user effort, participants spent between 6-10 minutes on average performing the splitting task, with the CARS dataset requiring the most time. The number of decisions made during the process ranged from approximately 95 to 120 across the datasets.\\
These results provide strong empirical validation of \method's effectiveness in real-world scenarios. The high agreement rates between users and our automated heuristics, particularly for CUB and CARS datasets, confirm that the proposed approach to identify inconsistent prototypes aligns well with human perception. Even with the more challenging PETS dataset, the method still achieved significant improvements, highlighting the robustness of \method across varying domains with different concept complexity levels.

\begin{table}[h!]
\caption{Results from prototype splitting across datasets, showing the percentage of prototypes identified as inconsistent before splitting and the percentage rated as more consistent after splitting. \textbf{Agreement} columns measure concept labeling consistency between different users (U.vs U.) and between users and our automated heuristic approach (U.vs H.). The \textbf{Time} column reports the average duration participants spent completing the splitting task, while \textbf{No. Decisions} indicates the average number of decisions made by participants.}
\centering
\begin{tabular}{lcccccc}
\toprule
\textbf{Dataset} & \textbf{Inconsistent} & \textbf{More consistent} & \multicolumn{2}{c}{\textbf{Agreement}} & \textbf{Time} & \textbf{No. Decisions} \\[-0.8em]
 &  &  & \multicolumn{2}{c}{\hrulefill} &  & \\
 & \textbf{Prototypes} & \textbf{After Split} & \textbf{U.vs U.} & \textbf{U. vs H.} & \textbf{(minutes)} & \\
\midrule
CUB & 91.7\% $\pm$ 9.0\% & 88.3\% $\pm$ 6.9\% & 89\% & 92\% & 6:09 & 120.4 \\
CARS & 86.1\% $\pm$ 5.9\% & 84.2\% $\pm$ 4.4\% & 87\% & 78\% & 9:45 & 113.2 \\
PETS & 70.6\% $\pm$ 24.6\% & 80.4\% $\pm$ 15.7\% & 50\% & 54\% & 7:58 & 95.2 \\
\bottomrule
\end{tabular}
\label{tab:user_study}
\vspace{-1.5em}
\end{table}

\section{Conclusion}\label{sec:conclusion}
In this work, we present \method, an interactive machine learning method designed to perform concept alignment. It is the first step towards personalized interpretability. \method shows that user feedback can be used not only to remove unwanted concepts and confounders but also to enhance the readability and understanding of explanations according to user preferences.

\paragraph{Limitations.}\label{sec:limitations}
Our current implementation of \method does not address cases where a prototypical part consists of more than two concepts. Although this limitation could be addressed by allowing the user to split prototypes in rounds, each time selecting the main two concepts until obtaining the desired consistency of the interpretations. Additionally, we do not explore other concept selection strategies from the user, as we found simple binary labeling to be both powerful and effective.
\paragraph{Impact.}
Our work lays the foundations for personalized interpretability, allowing users to adjust models to their specific needs through a few simple steps, resulting in well-understood, concept-level explanations. Moreover, we advocate for a broader view of model debugging, that is not only about removal of unwanted concepts and confunders, but also splitting information within models using user feedback. Such approach to model debugging not only enhances interpretability but also empowers users to actively participate in refining model outputs.

\paragraph{Future works.}
We see several directions to work on in the future. First, we aim to further personalize interpretations, leading to systems tailored to individual users. This can be achieved by studying user input types and multiple modalities to explain the concepts. Second, we intend to adapt this methodology to multimodal scenarios, particularly focusing on popular and large-scale visual-language models. Lastly, we will investigate the degree of personalization required for explanations across different data types, such as medical imaging, fine-grained classification, and natural images, to better understand user needs and expectations.

\begin{ack}
The work of T. Michalski was funded by the National Science Centre (Poland) grant no. 2022/47/B/ST6/03397.

The work of A. Wróbel, B. Zieliński and D. Rymarczyk was funded by "Interpretable and Interactive Multimodal Retrieval in Drug Discovery" project. The „Interpretable and Interactive Multimodal Retrieval in Drug Discovery” project (FENG.02.02-IP.05-0040/23) is carried out within the First Team programme of the Foundation for Polish Science co-financed by the European Union under the European Funds for Smart Economy 2021-2027 (FENG). 

A. Bontempelli acknowledges the support of the MUR PNRR project FAIR - Future AI Research (PE00000013) funded by the NextGenerationEU.

We gratefully acknowledge Polish high-performance computing infrastructure PLGrid (HPC Centers: ACK Cyfronet AGH) for providing computer facilities and support within computational grant no. PLG/2025/017993.

Some experiments were performed on servers purchased with funds from the Priority Research Area (Artificial Intelligence Computing Center Core Facility) under the Strategic Programme Excellence Initiative at Jagiellonian University.

Funded by the European Union. Views and opinions expressed are, however, those of the author(s) only and do not necessarily reflect those of the European Union or the European Health and Digital Executive Agency (HaDEA). Neither the European Union nor the granting authority can be held responsible for them. Grant Agreement no. 101120763 - TANGO.

\end{ack}

\newpage
\bibliographystyle{unsrtnat}
\bibliography{references}

\clearpage
\appendix
\section*{Supplementary Material}

\renewcommand{\thesection}{\Alph{section}}
\renewcommand{\thesubsection}{\thesection.\arabic{subsection}}

\section{Datasets}
We use 4 datasets in our paper:
\paragraph{FunnyBirds}
The FunnyBirds dataset \cite{hesse2023funnybirds} is a synthetic collection specifically designed for interpretability research in fine-grained classification. It consists of 50 classes constructed from combinations of 26 distinct bird parts. Each bird image is composed of anatomical components (head, beak, wings, etc.) with specific colors and shapes. The dataset includes 50,000 training images and 500 test images. A key feature of FunnyBirds is the inclusion of precise segmentation masks for all anatomical parts, which enables objective evaluation of prototype consistency and supports automated concept identification.
\paragraph{CUB}
The Caltech-UCSD Birds-200-2011 (CUB) dataset \cite{welinder2010caltech} contains 11788 images of 200 bird species. We follow the standard split with 5994 training images and 5794 test images. The dataset includes point annotations for 15 body parts per image, which we leverage to compute the purity metric.
\paragraph{CARS}
The Stanford Cars dataset \cite{krause20133d} contains 16185 images of 196 car classes. We use the standard split with 8144 training images and 8041 test images.
\paragraph{PETS}
The Oxford-IIIT Pet dataset \cite{parkhi2012cats} consists of 7390 images, out of which 3686 contain bounding boxes, of 37 pet categories (25 dog breeds and 12 cat breeds). We used those 3686 images that we split into 2953 and 733 for training and 733 testing, respectively.

\section{Computational Resources}
Training of baseline models required up to 40GB of VRAM and was performed using NVIDIA A100 GPUs, while the method development was conducted on NVIDIA RTX 3090 GPUs. We used a single GPU for approximately 5 hours to train the initial PIP-Net models on each dataset. The prototype splitting procedure required significantly less computational power and VRAM due to freezing most of the architecture. During the splitting process, only the weights corresponding to the inconsistent prototype and its duplicate are optimized, while the rest of the network remains fixed. Consequently, the resources needed to split the selected 10 prototypical parts per model could be performed under one hour per model on the NVIDIA RTX 3090, requiring less than 8GB of VRAM. This includes the computational cost of both the automatic selection of inconsistent prototypical parts and the identification of their two main concepts for splitting. The user study implementation, however, demanded greater computational resources as the splitting operations needed to be performed in real-time before users completed their labeling of all prototypical parts. To accommodate these requirements, we employed computing instances equipped with 8x NVIDIA RTX 4090 GPUs and limited concurrent participation to a maximum of two users to ensure optimal performance and responsiveness during the interactive sessions. Our software environment consisted of Python 3.9.19 with PyTorch 1.13.1 and CUDA 11.3 for model development and execution. The user study web interface was implemented using Flask 2.0.1 for the backend server and Node.js v20.18.0 for the frontend components. This research project's total consumed computational resources, including preliminary experiments and approaches not reported in the paper, amounted to approximately 1,800 GPU hours.

\section{Code}
Our implementation builds upon the original PIP-Net codebase by Nauta et al. \cite{nauta2023pip}. The original code is available at \url{https://github.com/M-Nauta/PIPNet}. We have properly credited the original authors in our code's documentation. Our modifications and extensions maintain compatibility.

\section{New Assets Documentation}
The code accompanying this paper is thoroughly documented to enable the reproducibility of our results. The repository includes:
\begin{enumerate}
\item \textbf{README.md}: Contains detailed setup instructions, dependency information, and a step-by-step guide to reproduce our experimental results.
\item \textbf{Method Documentation}: Comprehensive documentation of the \method method, including the prototype splitting mechanism and heuristics for detecting inconsistent prototypes.

\item \textbf{Configuration Files}: All hyperparameters and configuration settings used in our experiments.

\item \textbf{Preprocessing Scripts}: Code for data preparation and preprocessing for each dataset.

\end{enumerate}
The code is released under the MIT License, allowing for both academic and commercial use with proper attribution. No personal data from user studies is included in our released assets.

\section{User Study and Participant Compensation}
Our user studies were conducted via the Prolific platform with 54 participants (18 per dataset). We obtained informed consent from all participants at the beginning of the study. The consent form explicitly stated the purpose of the research, the tasks involved, and how the data would be used. No personally identifiable information was collected during the studies.
We compensated participants with the recommended hourly rate of £12.84/h, in accordance with Prolific's guidelines. The average completion time for the studies was 6 to 10 minutes, resulting in compensation of approximately £1.28 to £2.14 per participant. All participants were aged 18 or older, with a median age of 34.5 years.

\newpage

\section{Finding Optimal Similarity Threshold}

\begin{algorithm}[ht]
\caption{Finding optimal similarity threshold and concepts for prototype inconsistency detection.}
\label{alg:find_optimal_threshold}
\KwIn{$P_d = \{P_1, P_2, \ldots, P_D\}$ where $P_d$ consists of representations of patches highly activated for prototype d, threshold range $[\delta_{\min}, \delta_{\max}]$, step size $\delta_{step}$, minimum clique size $Q$}
\KwOut{Optimal threshold $\delta^*$, inconsistent prototypes with their cliques, split concepts}
Initialize $\text{best\_score} \leftarrow 0$, $\delta^* \leftarrow \delta_{\min}$\;

\tcp{Find optimal threshold by evaluating score($\delta$) = $\sum_{d=1}^{D} \mathbb{I}(C_d^1, C_d^2) \cdot \text{dissim}(C_d^1, C_d^2)$}
\For{$\delta = \delta_{\min}$ \KwTo $\delta_{\max}$ \textbf{by} $\delta_{step}$}{
    $\text{total\_dissim} \leftarrow 0$\;
    
    \For{$d = 1$ \KwTo $D$}{
        Construct similarity graph $G_d$ with edges where $\text{sim}(\mathbf{f}_i, \mathbf{f}_j) > \delta$\;
        Find cliques $\mathcal{C}_d = \{C_d^1, C_d^2, \ldots\}$ using Bron-Kerbosch algorithm and sort by size\;
        
        \If{$|\mathcal{C}_d| \geq 2$ \textbf{and} $|C_d^1| \geq Q$ \textbf{and} $|C_d^2| \geq Q$ \textbf{and} $C_d^1 \cap C_d^2 = \emptyset$}{
            $\text{dissim} \leftarrow 1 - \max\limits_{i \in C_d^1, j \in C_d^2} \text{sim}(\mathbf{f}_i, \mathbf{f}_j)$\;
            $\text{total\_dissim} \leftarrow \text{total\_dissim} + \text{dissim}$\;
        }
    }
    
    \tcp{Update optimal threshold if current score is better}
    $\text{current\_score} \leftarrow \text{total\_dissim}$\;
    \If{$\text{current\_score} > \text{best\_score}$}{
        $\text{best\_score} \leftarrow \text{current\_score}$\;
        $\delta^* \leftarrow \delta$\;
    }
}

\tcp{Identify and collect inconsistent prototypes using optimal threshold}
$\text{inconsistent\_protos} \leftarrow \{\}$\;
\For{$d = 1$ \KwTo $D$}{
    Construct similarity graph $G_d$ with edges where $\text{sim}(\mathbf{f}_i, \mathbf{f}_j) > \delta^*$\;
    Find = cliques $\mathcal{C}_d = \{C_d^1, C_d^2, \ldots\}$ using Bron-Kerbosch algorithm and sort by size\;
    
    \If{$|\mathcal{C}_d| \geq 2$ \textbf{and} $|C_d^1| \geq Q$ \textbf{and} $|C_d^2| \geq Q$ \textbf{and} $C_d^1 \cap C_d^2 = \emptyset$}{
        $\text{dissim}_d \leftarrow 1 - \max\limits_{i \in C_d^1, j \in C_d^2} \text{sim}(\mathbf{f}_i, \mathbf{f}_j)$\;
        $\text{inconsistent\_protos} \leftarrow \text{inconsistent\_protos} \cup \{(d, \text{dissim}_d, C_d^1, C_d^2)\}$\;
    }
}

\tcp{Sort prototypes by internal dissimilarity (highest first)}
Sort $\text{inconsistent\_protos}$ in descending order by $\text{dissim}_d$\;

\tcp{Split prototypes starting from highest internal dissimilarity}
$\text{split\_concepts} \leftarrow \{\}$\;
\For{\textbf{each} $(d, \text{dissim}_d, C_d^1, C_d^2) \in \text{inconsistent\_protos}$}{
    Split prototype $d$ into two concepts:\;
    \quad Concept 1: patches with indexes $C_d^1$\;
    \quad Concept 2: patches with indexes $C_d^2$\;
    $\text{split\_concepts} \leftarrow \text{split\_concepts} \cup \{(d, \text{Concept 1: } C_d^1, \text{Concept 2: } C_d^2)\}$\;
}

\Return $\delta^*$, $\text{inconsistent\_protos}$, $\text{split\_concepts}$\;
\end{algorithm}

The proposed heuristics not only detect inconsistent prototypical parts but also assign their patches to two major concepts. To this end, it requires selecting an optimal similarity threshold $\delta$ that identifies prototypes containing distinct clusters in feature space, as shown in the~\Cref{alg:find_optimal_threshold}. 
Given the feature representation $f(\textbf{x}) \in \mathbb{R}^{H \times W \times D}$, we analyze the activation patterns across patches for each prototype $d \in \{1,\ldots,D\}$.
To obtain the set of patches $P_d$ with high activation for this prototype. Each patch is represented by its corresponding feature vector $\mathbf{f}_i \in \mathbb{R}^D$ in the feature space. We construct a similarity graph $G_d = (V_d, E_d)$ where:
\begin{itemize}
    \item Each vertex $v_i \in V_d$ represents a feature vector $\mathbf{f}_i$ from $P_d$
    \item An edge $(i,j) \in E_d$ exists if and only if the cosine similarity between features exceeds our threshold: $\text{sim}(\mathbf{f}_i, \mathbf{f}_j) > \delta$
\end{itemize}
The algorithm evaluates multiple candidate thresholds $\delta \in [\delta_{\min}, \delta_{\max}]$ to identify the optimal value $\delta^*$ that maximizes:
$$\text{score}(\delta) = \sum_{d=1}^{D} \mathbb{I}(C_d^1, C_d^2) \cdot \text{dissim}(C_d^1, C_d^2)$$
Where:
\begin{itemize}
    \item $C_d^1, C_d^2$ represent the two largest disjoint cliques in $G_d$ identified using the Bron-Kerbosch algorithm
    \item $\mathbb{I}(C_d^1, C_d^2)$ is an indicator function that equals 1 if prototype $d$ contains two disjoint cliques of size at least $Q$, and 0 otherwise
    \item $\text{dissim}(C_d^1, C_d^2) = 1 - \max\limits_{i \in C_d^1, j \in C_d^2} \text{sim}(\mathbf{f}_i, \mathbf{f}_j)$ measures the dissimilarity between the most similar elements of both cliques.
\end{itemize}
This formulation identifies prototypes containing the most clearly separable concepts, which become candidates for splitting. Finally, the prototypes are then sorted according to their $\text{dissim}(C_d^1, C_d^2)$ and split starting from the prototypical parts with the highest internal dissimilarity.

\section{Examples of automatic splitting}
\begin{figure}
    \centering
    \includegraphics[width=0.8\linewidth]{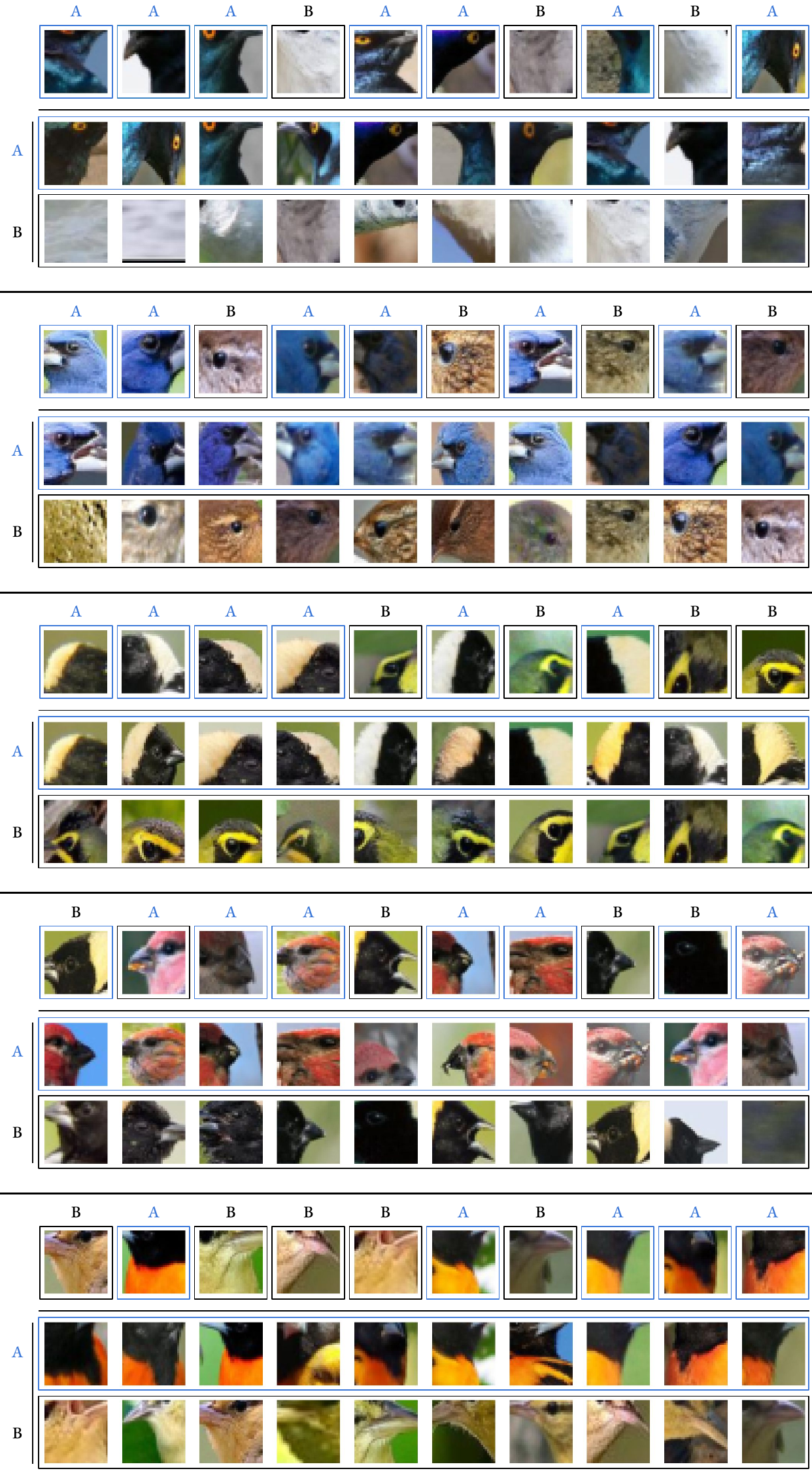}
    \caption{Automatic selection of concepts (A/B) and splitting of the first 5 out of 10 most inconsistent prototypical parts from the CUB dataset.}
    \label{fig:first_5of10}
\end{figure}

\begin{figure}
    \centering
    \includegraphics[width=0.8\linewidth]{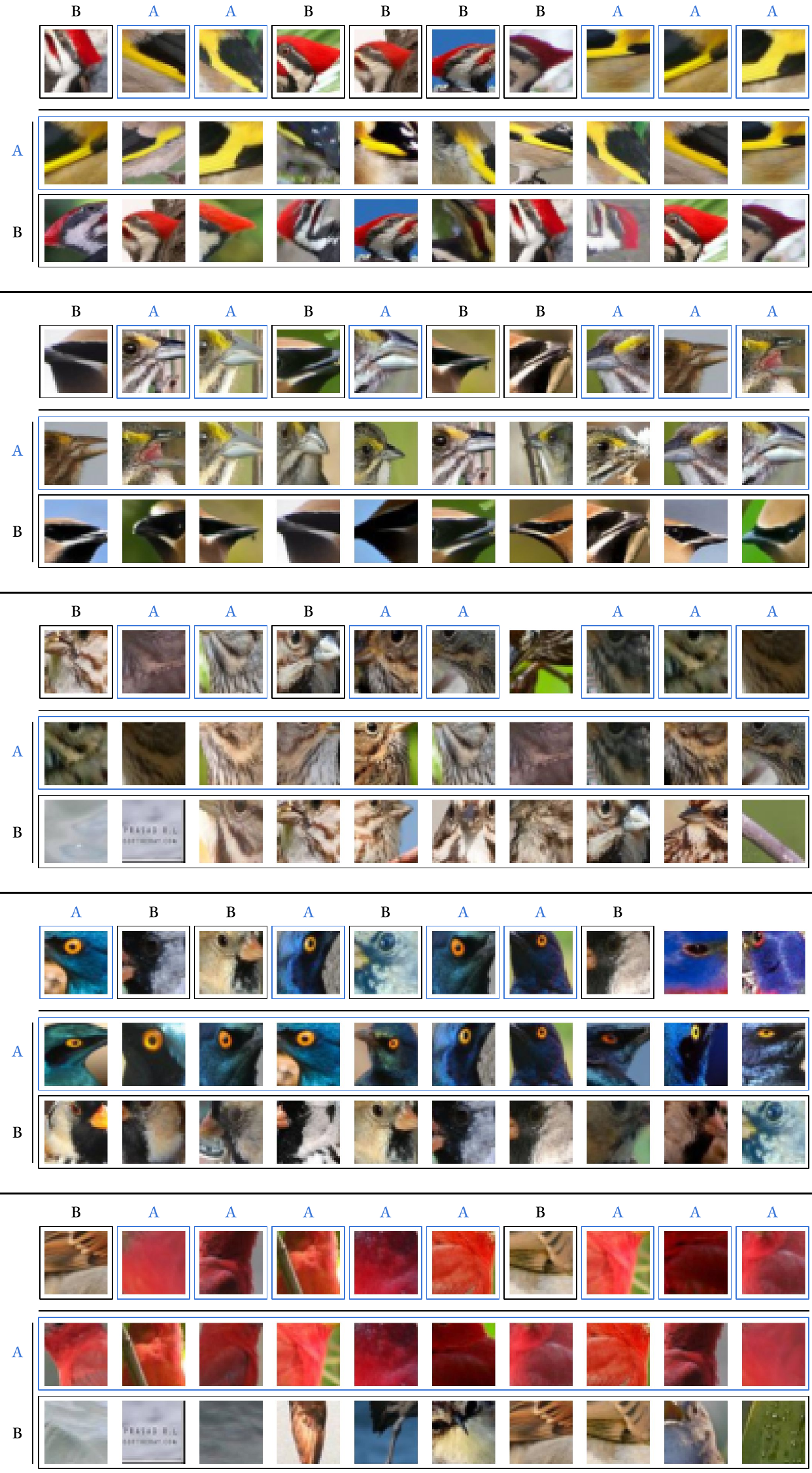}
    \caption{Automatic selection of concepts (A/B) and splitting of the remaining 5 out of 10 most inconsistent prototypical parts from the CUB dataset.}
    \label{fig:remaining_5o10}
\end{figure}

\Cref{fig:first_5of10,fig:remaining_5o10} present automatic splitting of the 10 most inconsistent prototypical parts from the CUB dataset.

\section{User Study}
\begin{figure}[h!]
    \centering
    \includegraphics[width=0.8\textwidth]{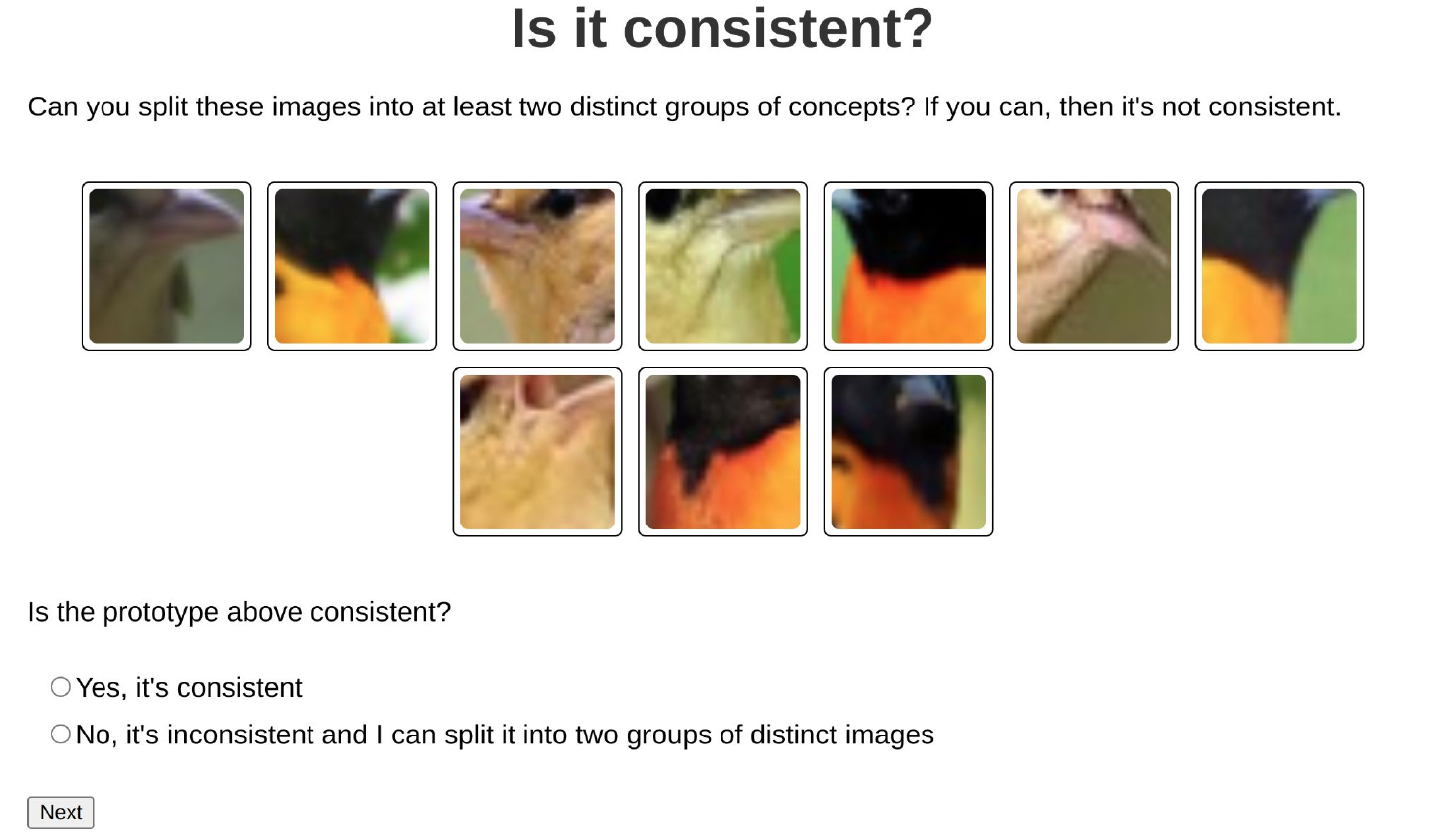}
    \caption{Phase I - user is asked if a prototypical part is consistent.}
    \label{fig:user_study_phase_1}
\end{figure}

\begin{figure}[h!]
    \centering
    \includegraphics[width=0.8\textwidth]{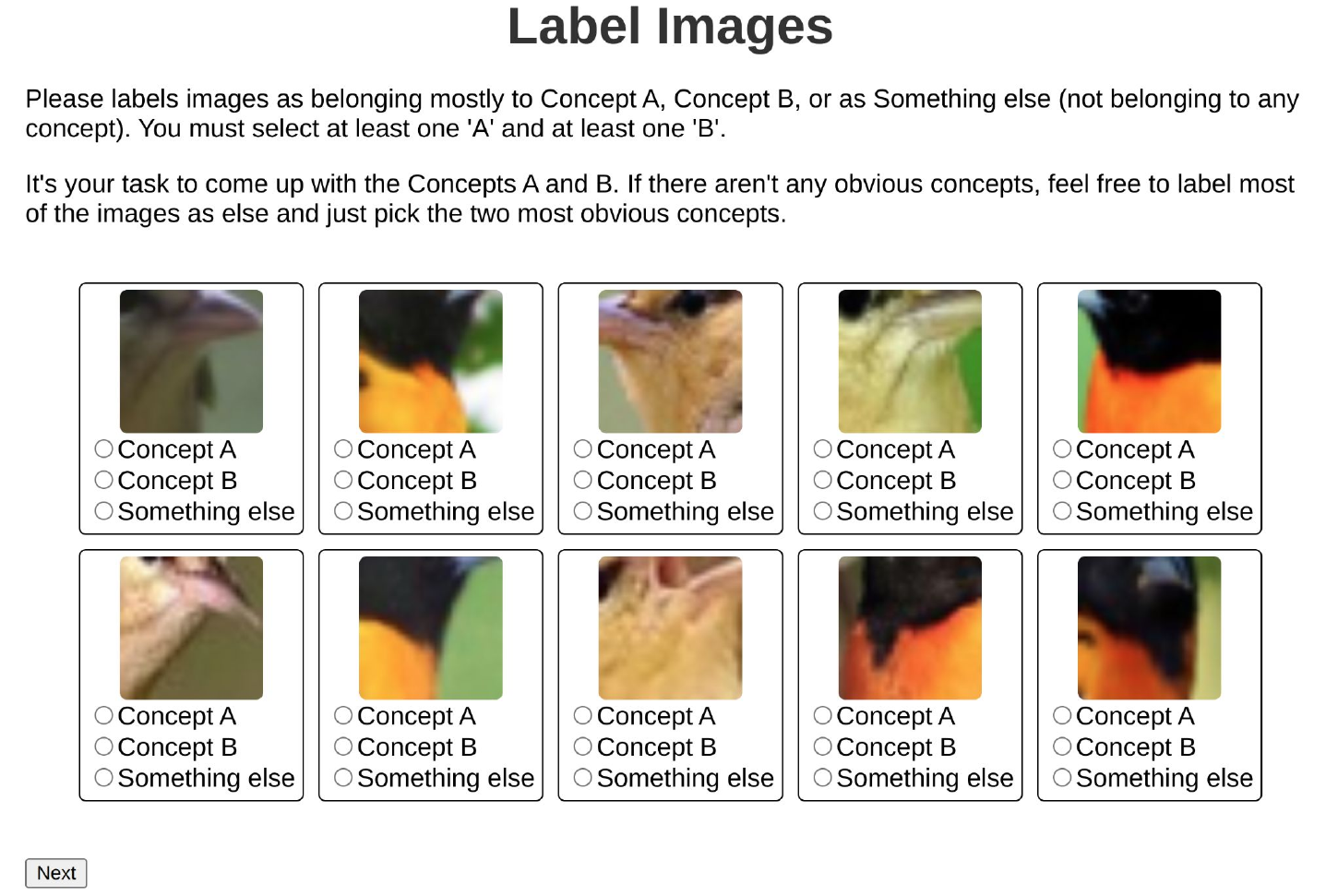}
    \caption{Phase II - user is asked to label patches as either Concept A, Concept B, or Something else.}
    \label{fig:user_study_phase_2}
\end{figure}

\begin{figure}[h!]
    \centering
    \includegraphics[width=0.8\textwidth]{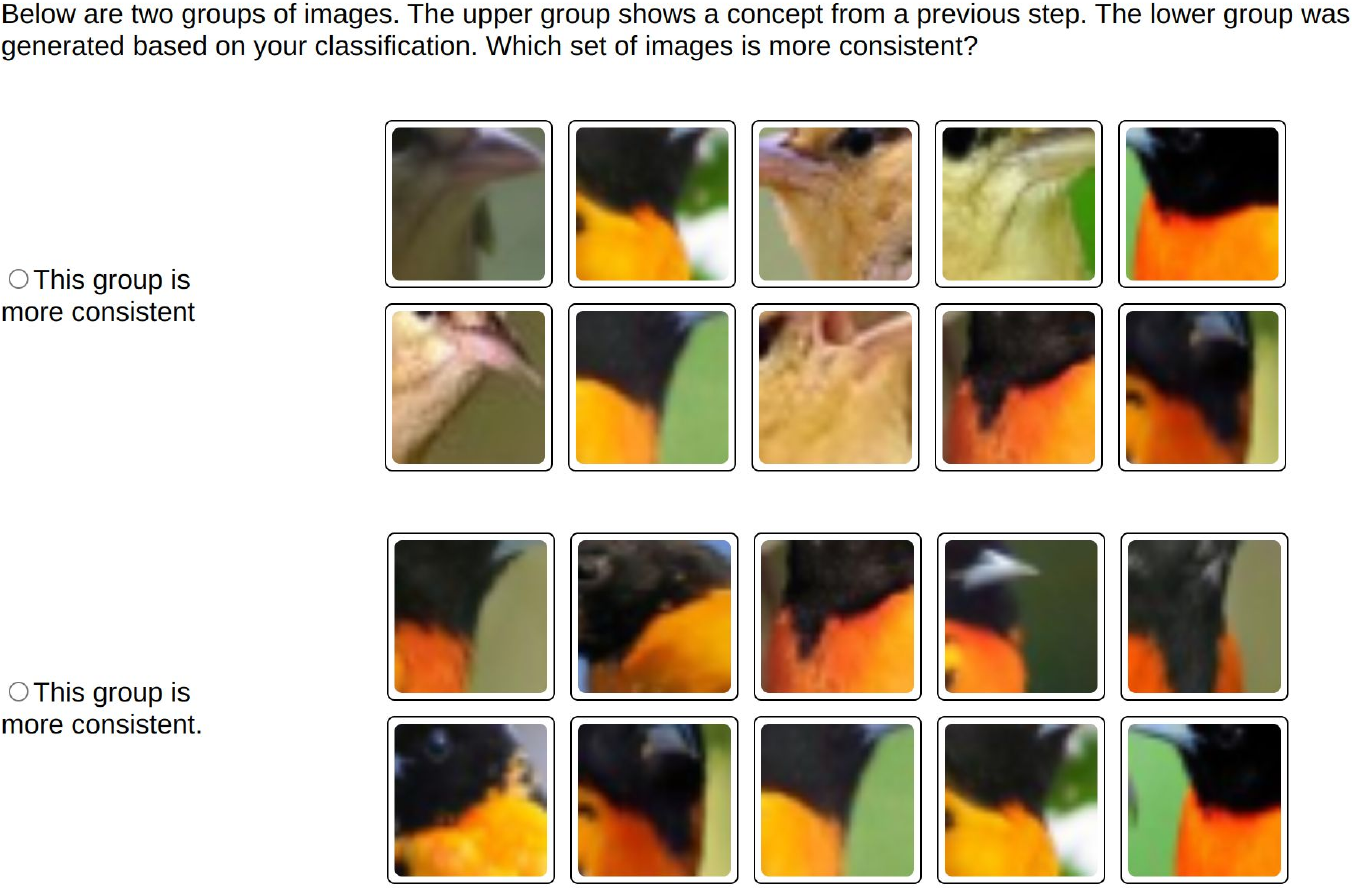} \\
\vspace{40pt}
    \includegraphics[width=0.8\textwidth]{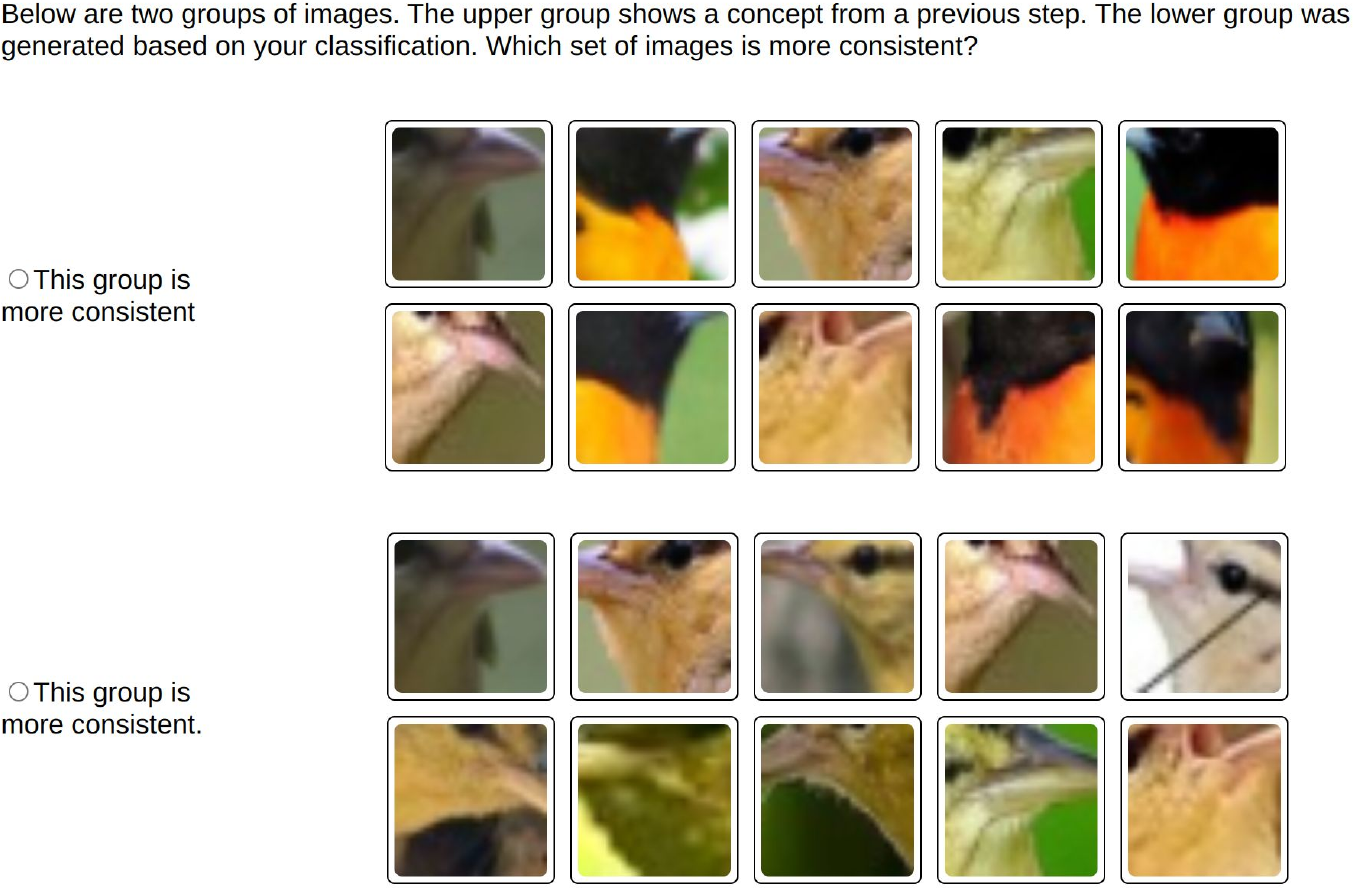}
    \caption{Phase III – user is asked if the split prototypical part is more consistent than before. Note that there are two prototypes created out of the single inconsistent one, and therefore the process requires two assessments.}
    \label{fig:user_study_phase_3_combined}
\end{figure}

\Cref{fig:user_study_phase_1,fig:user_study_phase_2,fig:user_study_phase_3_combined} present three phases of the user study. In phase I, users were asked if a proposed prototypical part was inconsistent. If they decided it was, they were redirected to phase II of this prototype, where they had to label each prototype's patches as one of the two concepts (A or B). Finally, after providing feedback for all of the prototypes, users were redirected to the III phase of the user study. At this point, they had to decide if the newly created prototype was more consistent than the original one. Note that at this stage, the feedback consists of two answers, one for each prototype that results from splitting the original one.

\vspace{40pt}

\end{document}